\documentclass{ecai} 
\usepackage{latexsym}
\usepackage{amssymb}
\usepackage{amsmath}
\usepackage{amsthm}
\usepackage{booktabs}
\usepackage{enumitem}
\usepackage{graphicx}
\usepackage{color, soul}
\usepackage{booktabs}
\usepackage{colortbl}
\usepackage{tcolorbox}
\usepackage{color,xcolor}
\usepackage{microtype}
\usepackage{inconsolata}
\usepackage{fancyvrb}
\usepackage{multirow}
\usepackage{lipsum}
\usepackage{pgfplots}
\usepackage{siunitx}
\usepackage{hyperref}
\usepackage{subcaption}
\usepackage{algorithm}
\usepackage{algpseudocode}
\pgfplotsset{compat=1.8}
\usepackage{hyperref}
\hypersetup{
    colorlinks=true,
    linkcolor=black,
    filecolor=black,
    urlcolor=black,
    citecolor=black,
}

\usepgfplotslibrary{statistics}

\newtheorem{theorem}{Theorem}

\newtheorem{definition}{Definition}

\newcommand{\BibTeX}{B\kern-.05em{\sc i\kern-.025em b}\kern-.08em\TeX}

\begin{document}

%%%%%%%%%%%%%%%%%%%%%%%%%%%%%%%%%%%%%%%%%%%%%%%%%%%%%%%%%%%%%%%%%%%%%%%%

\begin{frontmatter}

%%% Use this command to specify your submission number.
%%% In doubleblind mode, it will be printed on the first page.

\paperid{123} 

\title{Sub-SA: Strengthen In-context Learning via Submodular Selective Annotation}

\author[a]{\fnms{Jian}~\snm{Qian} \thanks{Corresponding Author. Email:jqian20@fudan.edu.cn.}\footnote{Equal contribution.}\label{fn:shared-footnote}}
\author[b]{\fnms{Miao}~\snm{Sun}}
\author[c]{\fnms{Sifan}~\snm{Zhou}\footnotemark[1]}
\author[c]{\fnms{Ziyu}~\snm{Zhao}}
\author[a]{\fnms{Ruizhi}~\snm{Hun}}
\author[a]{\fnms{Patrick}~\snm{Chiang}} 

\address[a]{State Key Laboratory of ASIC and System, Fudan University, Shanghai, China}
\address[b]{School of Electrical and Electronic Engineering, Nanyang Technological University, Singapore}
\address[c]{School of Automation, Southeast University, Nanjing, China}

\begin{abstract}
In-context learning (ICL) leverages in-context examples as prompts for the predictions of Large Language Models (LLMs). These prompts play a crucial role in achieving strong performance. However, the selection of suitable prompts from a large pool of labeled examples often entails significant annotation costs. To address this challenge, we propose \textbf{Sub-SA} (\textbf{Sub}modular \textbf{S}elective \textbf{A}nnotation), a submodule-based selective annotation method. The aim of Sub-SA is to reduce annotation costs while improving the quality of in-context examples and minimizing the time consumption of the selection process. In Sub-SA, we design a submodular function that facilitates effective subset selection for annotation and demonstrates the characteristics of monotonically and submodularity from the theoretical perspective. Specifically, we propose \textbf{RPR} (\textbf{R}eward and \textbf{P}enalty \textbf{R}egularization) to better balance the diversity and representativeness of the unlabeled
dataset attributed to a reward term and a penalty term, respectively. Consequently, the selection for annotations can be effectively addressed with a simple yet effective greedy search algorithm based on the submodular function. Finally, we apply the similarity prompt retrieval to get the examples for ICL. Compared to existing selective annotation approaches, Sub-SA offers two main advantages. \textbf{(1.)} Sub-SA operates in an end-to-end, unsupervised manner, and significantly reduces the time consumption of the selection process (from hours-level to millisecond-level). \textbf{(2.)} Sub-SA enables a better balance between data diversity and representativeness and obtains state-of-the-art performance. Meanwhile, the theoretical support guarantees their reliability and scalability in practical scenarios. Extensive experiments conducted on diverse models and datasets demonstrate the superiority of Sub-SA over previous methods, achieving millisecond(ms)-level time selection and remarkable performance gains. The efficiency and effectiveness of Sub-SA make it highly suitable for real-world ICL scenarios.
Our codes are available at \href{https://github.com/JamesQian11/SubSA}{https://github.com/JamesQian11/SubSA}

\end{abstract}

\end{frontmatter}

% 修改时态， 统一缩写
% 贡献 总结 abs写好
%%%%%%%%%%%%%%%%%%%%%%%%%%%%%%%%%%%%%%%%%%%%%%%%%%%%%%%%%%%%%%%%%%%%%%%%% 
\section{Introduction}
\label{Introduction}
Large language models (LLMs) can rapidly adapt to various downstream tasks from just a few examples provided in the prompt without parameter update~\cite{akyurek2022learning,bhattamishra2023understanding,brown2020language} while improving accuracy~\cite{liu2021makes, yoo2022ground}.
This remarkable ability is called in-context learning (ICL). 
Unlike traditional methods ~\cite{liu2021swin, carion2020end, shan2021ptt} that rely on task-specific data to update model parameters, ICL harnesses the powerful few-shot learning ability of LLMs effectively, making it a promising application.
Recent works~\cite{liu2021makes,rubin2021learning,pryzant2023automatic} show that collecting prompts from a broad selection of annotated examples is vital to achieving high-quality performance. Specifically, these papers have indicated that the performance substantially improves when similar examples (with some embedding criteria) are retrieved as in-context examples for the single test case. Given the varying requirements of different test cases for tailored in-context examples, the availability of a comprehensive collection of annotated examples assumes utmost importance in boosting the performance of LLMs.

\definecolor{color5}{RGB}{214, 39, 40} % Random
\definecolor{color6}{RGB}{31, 119, 180} % Vote-k
\definecolor{color7}{RGB}{44, 160, 44} % IDEAL
\definecolor{color8}{RGB}{255, 165, 0} % Vote
% 定义新的标记样式
\pgfdeclareplotmark{filledrectangle}{%
   \fill[shift={(0pt, 0pt)}] rectangle (-0pt, -0pt) rectangle (0pt, 0pt); % Adjust size here as needed
}

\begin{figure}[t]
\centering
\resizebox{0.9\columnwidth}{!}{%
\begin{tikzpicture}
\begin{axis}[
    xlabel={\textbf{Log} Time (ms)},
    ylabel={Performance (\%)},
    xmin=1, xmax=7.5,
    ymin=30, ymax=110,
    grid=both,
    minor tick num=1,
    major grid style={line width=.2pt,draw=gray!50},
    minor grid style={line width=.1pt,draw=gray!25},
    legend style={at={(0.99,0.99)}, 
    anchor=north east, 
    legend columns=1,
    font=\scriptsize},
    legend cell align=left,
    name=mainplot % 命名这个axis，便于之后引用位置
]
\addlegendimage{only marks, mark=filledrectangle, mark size=2.5pt, color=color6}
\addlegendentry{\textcolor{color6}{\textbf{Vote-$k$ (ICLR 2023)}}}
\addlegendimage{only marks, mark=filledrectangle, mark size=2.5pt, color=color7}
\addlegendentry{\textcolor{color7}{\textbf{IDEAL  (ICLR 2024)}}}
\addlegendimage{only marks, mark=filledrectangle, mark size=2.5pt, color=color5}
\addlegendentry{\textcolor{color5}{\textbf{Sub-SA (Ours)}}}
% \addlegendimage{only marks, mark=filledrectangle, mark size=2.5pt, color=color5}
% \addlegendentry{\textcolor{color5}{\textbf{Sub-SA (Ours)}}}
% Red points with different shapes
\addplot[only marks, mark=*, mark size=2.5pt, color=color6] coordinates {(6.676, 61.1)};
\addplot[only marks, mark=square*, mark size=2.5pt, color=color6] coordinates {(6.851, 39.5)};
\addplot[only marks, mark=triangle*, mark size=2.5pt, color=color6] coordinates {(6.677, 85.5)};
\addplot[only marks, mark=diamond*, mark size=2.5pt, color=color6] coordinates {(6.597, 57.2)};
\addplot[only marks, mark=pentagon*, mark size=2.5pt, color=color6] coordinates {(7.07, 40.0)};
% Blue point with a different shape
\addplot[only marks, mark=*, mark size=2.5pt, color=color7] coordinates {(5.916, 65.1)};
\addplot[only marks, mark=square*, mark size=2.5pt, color=color7] coordinates {(5.904, 39.5)};
\addplot[only marks, mark=triangle*, mark size=2.5pt, color=color7] coordinates {(5.918, 78.0)};
\addplot[only marks, mark=diamond*, mark size=2.5pt, color=color7] coordinates {(5.773, 57.3)};
\addplot[only marks, mark=pentagon*, mark size=2.5pt, color=color7] coordinates {(5.919, 45.8)};
% Green point with a different shape
\addplot[only marks, mark=*, mark size=2.5pt, color=color5] coordinates {(2.279, 65.9)} node[left, black, xshift=0pt] {\scriptsize{MRPC}} {};
\addplot[only marks, mark=square*, mark size=2.5pt, color=color5] coordinates {(2.342, 40.4)} node[left, black, xshift=0pt] {\scriptsize{MNLI}} ;
\addplot[only marks, mark=triangle*, mark size=2.5pt, color=color5] coordinates {(2.38, 90.6)} node[left, black, xshift=0pt] {\scriptsize{SST-2}};
\addplot[only marks, mark=diamond*, mark size=2.5pt, color=color5] coordinates {(2.30, 61.5)} node[left, black, xshift=0pt] {\scriptsize{RTE}};
\addplot[only marks, mark=pentagon*, mark size=2.5pt, color=color5] coordinates {(2.255, 49.7)} node[left, black, xshift=0pt] {\scriptsize{SST-5}};
% Green point with a different shape
\addplot[dashed, thick, color=lime] coordinates {(6.676, 61.1) (5.916, 65.1) (2.279, 65.9)};
\addplot[dashed, thick, color=magenta] coordinates {(6.851, 39.5) (5.904, 39.5) (2.342, 40.4)};
\addplot[dashed, thick, color=pink] coordinates {(6.677, 85.5) (5.918, 78.0) (2.38, 90.6)};
\addplot[dashed, thick, color=cyan] coordinates {(6.597, 57.2) (5.773, 57.3) (2.30, 61.5)};
\addplot[dashed, thick, color=yellow] coordinates {(7.07, 40.0) (5.919, 45.8) (2.255, 49.7)};
% Green point with a different shape
\addplot[violet, very thick, -stealth] coordinates {(6.677, 85.5) (2.38, 85.5)};
\addplot[violet, very thick, -stealth] coordinates {(2.38, 85.5) (2.38, 90.6)};
\node at (axis cs:4,85.5) [above right] {\textbf{\textcolor{violet}{\scriptsize{2.8x}}}};
\node at (axis cs:2.4,86) [above right] {\textbf{\textcolor{violet}{\scriptsize{+5.1}}}};
% Green point with a different shape
\addplot[orange, very thick, -stealth] coordinates {(5.918, 78.0) (2.38,78.0)};
\addplot[brown, very thick, -stealth] coordinates {(2.38,78.0) (2.38, 90.6)};
\node at (axis cs:3.5,78.0) [above right] {\textbf{\textcolor{orange}{\scriptsize{2.4x}}}};
\node at (axis cs:1.6,80.0) [above right] {\textbf{\textcolor{orange}{\scriptsize{+12.6}}}};
% (6.7742,56.66)
% (5.886,57.14)
% (2.3112,61.62)
\end{axis}
\end{tikzpicture}
}
\caption{Comparison on performance and time consumption during subset selection under the same hardware condition. The $y$-axis represents the performance of different datasets on the classification task, and the $x$-axis represents the time consumption with the \textbf{log} scale. Here the annotation budget is 18. Our Sub-SA can remarkably outperform the Vote-$k$ baseline by a large average margin ($5.0\%$ absolute gain and $2.9\times$ acceleration of millisecond-level log representation). Notably, Sub-SA outperforms the IDEAL on the SST-2 benchmark with an improvement of $12.6\%$ absolute gain.
}
\vspace{15pt}
\label{fig:PT}
\end{figure}
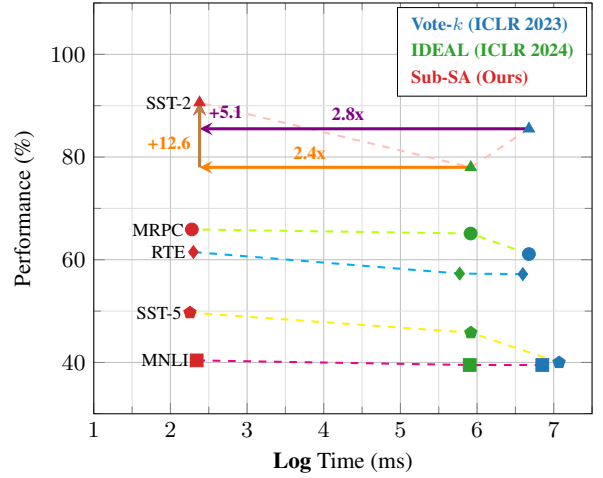

However, compiling and annotating extensive datasets to obtain corresponding examples for ICL demands substantial manpower and financial costs. For example, when an annotator labels a question and answer pair, they are also required to provide a rational, step-by-step thinking approach on how to obtain the final output~\cite{wei2022chain}, which significantly increases the annotation costs.

\begin{figure}[htbp]
\centering
\includegraphics[width=1\columnwidth]{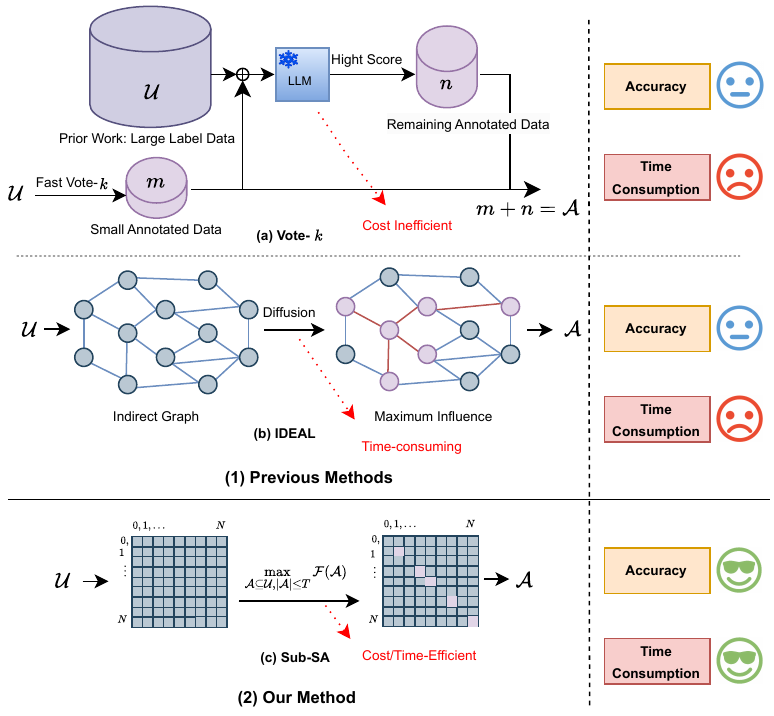}
\caption{Comparison on different methods for selective annotation.}
\vspace{15pt}
\label{fig: Methods}
\end{figure}

To address the challenges of annotation cost, prior researches endeavor Vote-$k$ and IDEAL~\cite{su2022selective,zhang2023ideal}. Vote-$k$~\cite{su2022selective} selects a $diverse$ and $representative$ subset from a large-scale unlabeled data pool for annotation based on the LLM's feedback to avoid retrieving all the examples. However, as shown in Figure~\ref{fig: Methods} (a) Vote-$k$ is subject to three drawbacks that limit its practical application. \textbf{(1.)} Vote-k places excessive emphasis on diversity in data selection, resulting in the neglect of representative data and leading to the selection of outliers~\cite{zhang2023ideal}. \textbf{(2.)} Vote-$k$ is not an end-to-end pipeline, leading to increased processing complexity and additional inference costs. \textbf{(3.)} Vote-$k$ lacks theoretical support to guarantee their reliability and scalability in practical scenarios. Following that, IDEAL~\cite{zhang2023ideal} utilizes an influence-driven mechanism to select annotation from a directed graph, which is built from the unlabeled dataset. Particularly, as shown in Figure~\ref{fig: Methods} (b), IDEAL sets every unlabeled data as a vertex and constructs the directed graph by considering the similarities between every example as the edges. After that, the influence of each candidate unlabeled subset is evaluated through an information diffusion process. Finally, a subset with high influence is iteratively selected based on a simple greedy algorithm. However, despite its performance well in empirical evaluations, IDEAL still lacks effectiveness in real-world scenarios with the following limitations.
\textbf{(1.)} IDEAL applies a classic independent-cascade diffusion model to select a subset, thereby ignoring the diversity and representativeness of unlabeled examples, resulting in the unsatisfactory performance of the method.
\textbf{(2.)} The process of examples selection from the graph, which involves repeating multiple steps of the diffusion process ten times, significantly increases computational complexity and renders the method highly inefficient.

In this paper, to minimize annotation costs for ICL and overcome the above issues of existing studies, an unsupervised, end-to-end submodule-based selective annotation method is proposed, called \textbf{Sub-SA} (\textbf{Sub}modular \textbf{S}elective \textbf{A}nnotation). In general, \textbf{Sub-SA} aims to identify a suitable subset of data from a large unlabeled dataset to serve as a proxy and a viable substitute for large annotated examples in subsequent ICL tasks. In detail, our Sub-SA has two core designs, which are submodular function and \textbf{RPR} (\textbf{R}eward and \textbf{P}enalty \textbf{R}egularization). Submodular function aim to facilitate efficient and effective subset selection from a large unlabeled dataset. Notably, due to the design attributes to the design of the submodular function, our selection process does not involve the LLMs~\cite{su2022selective} or iterative selection~\cite{zhang2023ideal}, thereby exponentially reducing the time-consuming of the selection process. RPR aims to better balance the diversity and representativeness of the unlabeled dataset and can be seamlessly integrated into the submodule selection function. Specifically, in RPR, the reward term encourages the selection of candidate elements with higher representative scores based on their similarity to the overall elements. On the other hand, the penalty term reduces the selection of candidate elements that are too similar to the elements already selected within the set. Subsequently, based on the results of the submodular function, a simple yet effective greedy search algorithm is utilized to obtain a subset of annotated examples. Finally, we apply similarity prompt retrieval to get the examples for the ICL task.

Theoretically, we demonstrate that our Sub-SA possesses the characteristics of monotonicity and submodularity, which provide the corresponding lower bound of the subset selection. Empirically, we validate the performance of our proposed approach over 8 datasets across diverse tasks (covering classification, commonsense reasoning, dialogue, and generation). Various LLMs (parameters from 2.7B to 175B) and prompt retrieval technologies are also included in evaluations. Comprehensive experiments show that the proposed method achieves further performance improvement compared to the state-of-the-art methods in 13 out of 16 cases. Meanwhile, it is worth noting that our method significantly reduces the time consumption of the selection process, from hours-level to millisecond-level, achieving a speedup of magnitudes around 10,000$\times$ (Table~\ref{con: time consumption}). In particular, with an annotation budget of either 18 or 100 from a 3K unlabeled dataset, the selective annotation cost decreases to 30-150$\times$. Figure~\ref{fig:PT} presents the comparison on performance and time consumption, our Sub-SA outperforms the Vote-$k$ and IDEAL methods, which makes it more suitable for real-world ICL scenarios. Furthermore, we compare different prompt retrieval methods for ICL tasks, our method consistently achieves better results than reference methods. This creates a strong baseline of selective annotations for follow-up research. 
% We will release our code to the community.

In summary, the main contributions of this paper are  as follows: 
\begin{itemize}
\item We propose a novel submodule-based selective annotation method named Sub-SA, which is an unsupervised, end-to-end subset selection technique for ICL. Furthermore, we theoretically demonstrate that our Sub-SA possesses the characteristics of monotonically and submodularity, which provide the corresponding lower bound of the subset selection.
\item We design RPR (Reward and Penalty Regularization) for Sub-SA, which consists of a reward term and a penalty term, to balance the diversity and representativeness of the unlabeled dataset selection.
\item Extensive experiments demonstrate that Sub-SA shows better performance than existing selective annotation methods across diverse tasks and models. More importantly, Sub-SA significantly reduces the time consumption during subset selection, from hours-level to millisecond-level, making it highly suitable for practical ICL scenarios. 
% We will release our code to the community.
\end{itemize}
%%%%%%%%%%%%%%%%%%%%%%%%%%%%%%%%%%%%%%%%%%%%%%%%%%%%%%%%%%%%%%%%%%%%%%%%% 
\section{Preliminaries}
\label{Preliminaries}

\subsection{In-context Learning}
In-context learning (ICL) is a pivotal capability behind the astounding performance of Large Language Models (LLMs). The combination of test input with a few input-output demonstrations allows these models to undertake a variety of tasks without updating parameters.
Formally, a $k$-shot prompt for ICL consists of $k$ examples, given a test sample $\left(x_{\text {test }}, y_{\text {test }}\right)$, LLMs predicts $\hat{y}$ based on the in-context prompts and input $x_{\text {test }}$ :
\begin{equation}
   \hat{y}=\operatorname{LLM}\left(e_k \oplus, \ldots, \oplus e_1 \oplus x_{\text {test }}\right) 
\end{equation}
where $e_i=\left(x_i, y_i\right)_{i=1}^k$ represents an sequence consisting of input-output pairs, $k$ denotes the shot number and $\oplus$ is the operation of concatenation. It is paramount to optimize the in-context prompts by seeking the ideal example set $\left\{e_k, \ldots, e_1\right\}$ in $\mathcal{A}$ for $x_{\text {test }}$, aiming to make the LLM's prediction $\hat{y}$ match the ground truth $y_{\text {test}}$, where $\mathcal{A}=\left\{\left(x_i, y_i\right)\right\}_{i=1}^T$  is the annotated examples and $T$ is the fixed budget.

\subsection{Submodular Subset Selection}
Considering a finite set $V$, given a set function $\mathcal{F}: 2^V \rightarrow {R}$ that maps any subset $A \subseteq V$ and $B \subseteq V$ to a real value. When $\mathcal{F}$ is a submodular function, its definition is as follows:
\begin{definition}[Submodular Function~\cite{edmonds2003submodular}]
For any set $A \subseteq B \subseteq V$. Given an element $\alpha$, where $\alpha=V \backslash B$. The set function $\mathcal{F}$ is a submodular function when it satisfies monotonically non-decreasing:
\begin{equation}
\mathcal{F}\left(A \cup\{\alpha\}\right)-\mathcal{F}\left(A\right) \geq 0
\end{equation}
And:
\begin{equation}
\mathcal{F}\left(A \cup\{\alpha\}\right)-\mathcal{F}\left(A\right) \geq \mathcal{F}\left(B \cup\{\alpha\}\right)-\mathcal{F}\left(B\right) .
\end{equation}
\label{df:submodularity}
\end{definition}
This definition indicates that the submodular function possesses a diminishing returns characteristic, allowing the subset selection problem to be framed as either minimizing or maximizing a submodular function~\cite{iyer2021submodular}.
It naturally measures the diversity and information of select set $A$, which ensures that each additional element contributes the maximum possible value to the overall objective. 

\begin{figure}[htbp]
\centering
\includegraphics[width=1\columnwidth]{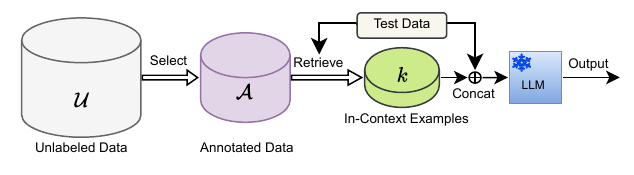}
\caption{The selective annotation pipeline for in-context learning. Given a pool of unlabeled instances $\mathcal{U}$, the goal of selective annotation is to select the most informative examples $\mathcal{A}$ to annotate based on a fixed budget $T$.Then, retrieve the annotated set based on unseen test data to get the $k$  shot examples, and finally concatenate the test data and $k$ in-context examples to maximize ICL performance with frozen LLM.}
\vspace{10pt}
\label{fig:Sub-SA}
\end{figure}

\subsection{Selective annotation Problem Setting}
As show in Figure~\ref{fig:Sub-SA}, given a pool of unlabeled samples $\mathcal{U}=\left\{x_i\right\}_{i=1}^N$, where $N$ represents the count of unlabeled instances. 
The goal of selective annotation is to select the most informative subset $\mathcal{A} \subset \mathcal{U}$ to make manual annotation, the size of $\mathcal{A}$ is the fixed budget $T$. After annotation, we can obtain the annotated dataset $\mathcal{A}=\left\{\left(x_i, y_i\right)\right\}_{i=1}^T$. Then, ICL applies prompts retrieve to get the $k$-shot ICL examples $e_k$ from the selected subset $\mathcal{A}$, which can perform on an undetected test set. 

In this paper, we concentrate on the selective annotation process to construct a new prompt that achieves efficient and effective in-context learning. With a monotonically non-decreasing submodular function $\mathcal{F}: 2^V \rightarrow \mathbb{R}$, the selective annotation problem can be viewed as maximizing the value $\mathcal{F}(\mathcal{A})$ with fixed budget $T$. Mathematically, the goal is to select a set $A$ consisting of a limited number $T$ examples through the  submodular function $\mathcal{F}$ in the large-scale unlabeled data $\mathcal{U}$:
\begin{equation}
\max _{\mathcal{A} \subseteq \mathcal{U},|\mathcal{A}| \leq T} \mathcal{F}(\mathcal{A})
\label{eq:max}
\end{equation}

% In this paper, we focus on selective annotation process by design a novel submodular function that enhances the diversity of the selected dataset while preserving the overall quality of similarity. 

%%%%%%%%%%%%%%%%%%%%%%%%%%%%%%%%%%%%%%%%%%%%%%%%%%%%%%%%%%%%%%%%%%%%%%%%% 
\section{Methodology}
\label{Methodology}
In this section, we formally introduce Sub-SA (Submodular Selective Annotation), an efficient and effective example selection approach for real-world in-context learning (ICL) scenarios. Figure~\ref{fig:Sub-SA} presents the overall pipeline of selection annotation for ICL in large language model inference. 
In Section~\ref{sub: selective annotation}, we introduce the process of Sub-SA, including 
the monotonous submodular function and a naive greedy algorithm to search the most informative subset with the maximum function. Figure~\ref{fig: Methods} (c) shows the framework of our Sub-SA. In Section~\ref{sub: prompt retrieval}, we introduce the prompt retrieval method to get the $k$ shot examples for LLM.

\subsection{Selective Annotation}
\label{sub: selective annotation}
For ICL prompt examples selections, it is cost-effective to identify an information-dense subset from a large-scale unlabeled data pool.  
Inspired by submodularity~\cite{iyer2021submodular}, which naturally evaluates the representative and diversity of the dataset.
We propose Sub-SA, which adopts a submodular function that consists of a reward term for representative scores and a penalty term for diverse scores to select the most informative examples within a fixed annotation budget for prompt retrieval.

% \subsubsection{Submodular Function}
\subsubsection{Reward and Penalty Regularization}
% 如何证明这个函数具有子模性质
In order to quantify the influence of all elements in the large-scale unlabeled dataset, we apply Sentence-BERT~\cite{reimers2019sentence} to calculate a vector representation $emb$. Then, we apply a non-negativity cosine similarity kernel $s$ to evaluate the example relationship. Finally, we introduce a submodular gain function with reward term $S_{\text {repres.}}$ and penalty term  $S_{\text {diverse.}}$to choose the most informative examples to annotate. 

\vspace{10pt} % 增加 12 磅的垂直间距
\noindent\textbf{Representative Score.} As previously mentioned, annotating large unlabeled data is cost-ineffective. Here we propose $S_{\text {repres.}} $, which characterizes the representation score of the large-scale dataset selection. 
\begin{equation}
S_{\text {repres.}}
= \sum_{i \in \mathcal{U}}  \sum_{k \in \mathcal{A}}\frac{emb[i] \cdot emb[k] }{\left\|emb[i]\right\| \cdot \left\| emb[k]\right\|}
\end{equation}
where $\mathcal{U}$ is the large-scale unlabeled samples, $\mathcal{A}$ is the selected subset of $\mathcal{U}$. $i$ is the elements of ublabeled dataset $\mathcal{U}$ and $k$ is the elements of dataset $\mathcal{A}$, respectively.

\vspace{10pt} % 增加 12 磅的垂直间距
\noindent\textbf{Diverse Score.}
For selective annotation, balancing diversity and representativeness is vital to improve the reliable and stable performance of LLM's prediction with ICL. 
Here we put forward $S_{\text {diverse.}}$, which calculates the diverse score between a selected subset of $\mathcal{U}$.
\begin{equation}
S_{\text {diverse.}}
=\sum_{m \in \mathcal{A}} \sum_{n \in \mathcal{A}} \frac{emb[m] \cdot emb[n] }{\left\|emb[m]\right\| \cdot \left\| emb[n]\right\|}
\end{equation}
where the $\mathcal{A}$ is the subset of large-scale unlabeled dataset $\mathcal{U}$,  $m$ and $n$ are the elements of the dataset $\mathcal{A}$.

\vspace{10pt} % 增加 12 磅的垂直间距
% \noindent\textbf{Submodular Function.}
\subsubsection{Submodular Function}
We construct the objective function $\mathcal{F}(\mathcal{A})$ for selective annotation through a combination of the above 
representative score and diverse score. 
\begin{equation}
\begin{aligned}
\mathcal{F}( \mathcal{A}) &= \lambda_1 S_{\text{repres.}} + \lambda_2 S_{\text{diverse.}} \\
&= \lambda_1  \sum_{i \in  \mathcal{U}} \sum_{k \in  \mathcal{A}} s_{ik} + \lambda_2 \sum_{m \in  \mathcal{A}} \sum_{n \in  \mathcal{A}} s_{m n}
\end{aligned}
\label{eq:gain1}
\end{equation}
where $\lambda_1$ and $\lambda_2$ represent the weighting factors used to trade-off between representativeness and diversity, $s$ is the similarity kernel. This time  $\lambda_1$ is set to 2 for representative dataset selection and  $\lambda_2$ is set to -1 for improving the diversity. We prove the submodularity and monotonically below.

\noindent\textbf{Proof.} This proof is inspired by~\cite{iyer2021submodular}. We first prove the monotonically, then we prove submodularity based on the first derivation process.

\noindent For monotonically proof, we have:
\begin{equation}
\centering
\resizebox{\columnwidth}{!}{$
\begin{aligned}
\mathcal{F}\left( \mathcal{A} \cup\{\alpha\}\right)&=\lambda_1 \sum_{i \in U} \sum_{k \in  \mathcal{A} \cup\{\alpha\}}   s_{i k}+\lambda_2 \sum_{m \in  \mathcal{A} \cup\{\alpha\}} \sum_{n \in  \mathcal{A} \cup\{\alpha\}} s_{m n}  \\
&= \lambda_1\left(\sum_{i \in U}\left(\sum_{k \in \mathcal{A}} s_{i k}+s_{i \alpha}\right)\right)+\lambda_2\left(\sum_{m \in \mathcal{A}} \sum_{n \in \mathcal{A}} s_{m n}+\sum_{m \in \mathcal{A}} s_{m \alpha}+\sum_{n \in \mathcal{A}} s_{n \alpha}+s_{\alpha \alpha}\right)
\end{aligned}$}
\end{equation}
Thus:
\begin{equation}
\centering
\resizebox{0.9\columnwidth}{!}{$
\mathcal{F}\left( \mathcal{A} \cup\{\alpha\}\right)-\mathcal{F}\left( \mathcal{A}\right)=\lambda_1 \sum_{i \in  \mathcal{U}} s_{\alpha i}+\lambda_2\left(2\sum_{m \in  \mathcal{A}} s_{m \alpha}+s_{\alpha \alpha}\right)
$}
\end{equation}
Since $\lambda_1 = 2$,  $\lambda_2 = -1$, and $s$ is non-negativity cosine similarity
kernel.
Follow up:
\begin{equation}
\centering
\resizebox{0.38\columnwidth}{!}{$
\mathcal{F}\left( \mathcal{A} \cup\{\alpha\}\right)-\mathcal{F}\left( \mathcal{A}\right)
\geq 0 $}
\end{equation}
% \mathcal{F}\left(A \cup\{\alpha\}\right)-\mathcal{F}\left(A\right)
% =2 \sum_{k \in U} s_{\alpha k}-\left(2\sum_{i \in A} s_{i \alpha}+s_{\alpha \alpha}\right)
% \geq 0 $}
\noindent For submodularity proof, We have:
\begin{equation}
\centering
\resizebox{0.9\columnwidth}{!}{$
\mathcal{F}\left( \mathcal{A} \cup\{\alpha\}\right)-\mathcal{F}\left( \mathcal{A}\right)=\lambda_1 \sum_{i \in \mathcal{U}} s_{\alpha i}+\lambda_2\left(2\sum_{m \in  \mathcal{A}} s_{m \alpha}+s_{\alpha \alpha}\right)
$}
\end{equation}
\begin{equation}
\centering
\resizebox{0.9\columnwidth}{!}{$
\mathcal{F}\left(\mathcal{B} \cup\{\alpha\}\right)-\mathcal{F}\left(\mathcal{B}\right)=\lambda_1 \sum_{i \in \mathcal{U}} s_{\alpha i}+\lambda_2\left(2\sum_{m \in \mathcal{B}} s_{m \alpha}+s_{\alpha \alpha}\right)
$}
\end{equation}
Since: $\mathcal{A} \subseteq \mathcal{B}$,
Therefore:
\begin{equation}
\centering
\resizebox{0.45\columnwidth}{!}{$
\sum_{m \in  \mathcal{A}} s_{m \alpha} \leq \sum_{m \in \mathcal{B}} s_{m \alpha}
$}
\end{equation}
We can get:
\begin{equation}
\centering
\resizebox{0.7\columnwidth}{!}{$
\mathcal{F}( \mathcal{A} \cup\{\alpha\})-\mathcal{F}( \mathcal{A}) \geq \mathcal{F}(\mathcal{B} \cup\{\alpha\})-\mathcal{F}(\mathcal{B})
$}
\end{equation}

\subsubsection{Greedy Search Algorithm}
\renewcommand{\algorithmicrequire}{\textbf{Input:}} 
\renewcommand{\algorithmicensure}{\textbf{Output:}} 
\begin{algorithm}
\caption{A greedy search based algorithm for selective annotation}
\begin{algorithmic}[1]
\Require The annotation budget $T$, Unlabeled data $\mathcal{U}$ with index.
\Ensure The set $\mathcal{A}$  that includes $T$ examples to annotate.
\State \textbf{Initialize} $\mathcal{A}_0 \leftarrow \emptyset, i=0$
\While{$i<T$}
    \State $\mathbf{a}_i \leftarrow \arg \max _{\mathbf{a}_i \in \mathbf{U} \backslash \mathcal{A}_i } \mathcal{F}(\mathcal{A}_i \cup\{\mathbf{a}_i\})$
    \State $\mathcal{A}_{i+1} \leftarrow \mathcal{A}_i \cup \mathbf{a}_i$
    \State $i \leftarrow i+1$
\EndWhile
\State \Return $\mathcal{A}$
\end{algorithmic}
\label{ag:ag1}
\end{algorithm}
\noindent Given a large-scale unlabeled dataset $\mathcal{U}$, we can apply Equation~\ref{eq:max} to search the diversity and
representativeness subset  by selecting $T$ elements that maximize the value of the submodular Function~\ref{eq:gain1}. This problem can be proficiently addressed by implementing a naive greedy algorithm.
Referring to related works, we apply Algorithm~\ref{ag:ag1} to optimize the output of the submodular function $\mathcal{F}(\mathcal{A})$. Based on the Definition~\ref{df:submodularity} we have proved that the Function~\ref{eq:gain1} is a submodular function. According to the theory of~\cite{nemhauser1978analysis} , we have:

\begin{theorem} $A$ denotes the solution obtained by the greedy search approach, and $A^*$ denotes the optimal solution. If $\mathcal{F}(\cdot)$ is a submodular function, then the solution $A$ has the following approximation guarantee:
\begin{equation}
\mathcal{F}(A) \geq\left(1-\frac{1}{\mathrm{e}}\right) \mathcal{F}\left(A^*\right)
\end{equation}
where $\mathrm{e}$ is the base of natural logarithm.
\end{theorem}

\subsection{Prompt Retrieval}
\label{sub: prompt retrieval}
Following the above submodular selective annotation, we receive a set of annotated examples $\mathcal{A} $. Then, we retrieve $k$ shot examples from 
the annotated set $\mathcal{A} $ to concatenate with the test input. Finally, the prompts are given to the LLM for relevant task prediction.
As previous research~\cite{zhang2023ideal,su2022selective}, we apply Sentence-BERT~\cite{reimers2019sentence} to  evaluate embeddings for all annotated examples 
and pick out the instances of each test case derived from the similarity between examples and test cases.

%%%%%%%%%%%%%%%%%%%%%%%%%%%%%%%%%%%%%%%%%%%%%%%%%%%%%%%%%%%%%%%%%%%%%%%%% 
\begin{table*}[htbp]
\centering
\resizebox{0.85\textwidth}{!}{%
\begin{tabular}{ccccccccccccccccc} % 将所有的 'l' 改为 'c' 来实现居中对齐
\hline
\multicolumn{2}{c}{\textbf{Method}} & \multirow{2}{*}{\textbf{Publication}}& \multicolumn{5}{c}{\textbf{Classification}}   \\
\cmidrule(lr){1-2} \cmidrule(lr){4-8} 
\multirow{2}{*}{\textbf{$|\mathcal{A}|$}} & \multirow{2}{*}{\textbf{Selection}} & & MRPC  & MNLI  & SST-2 & RTE & SST-5  \\
\cmidrule(lr){4-8} 
& & & \scriptsize{Mean/Max/Min} & \scriptsize{Mean/Max/Min} & \scriptsize{Mean/Max/Min} & \scriptsize{Mean/Max/Min} & \scriptsize{Mean/Max/Min} \\
\hline
100 & Random  & - &64.3/68.4/58.6   & 38.4/39.5/37.5 & 88.0/88.7/87.5 & 56.9/60.5/55.1 & 48.2/51.2/43.4 \\
100 & Vote-$k$~\cite{su2022selective}  & ICLR 2023 & 63.8/68.8/60.9 & 39.3/43.8/36.7  & 90.5/91.4/89.5 & 57.6/57.8/57.4 & 45.4/46.5/44.1   \\
100 & IDEAL~\cite{zhang2023ideal}  & ICLR 2024 & 66.7/69.1/64.8  & 39.7/41.4/38.7  & 89.1/90.6/87.1 & 57.3/59.0/55.9 & \textbf{51.0/55.5/48.4}  \\
\rowcolor{gray!50}
100 & Sub-SA & Ours & \textbf{68.1/69.9/66.0}  & \textbf{43.2/46.5/39.1}  & \textbf{91.1/93.0/88.3}& \textbf{58.1/58.2/57.8} & 48.0/51.6/43.8  \\
\hline
18 & Random  & - & 57.4/68.8/39.8  & 38.3/42.2/34.4  & 85.7/87.9/81.3 & 56.0/57.8/53.1 & 47.1/49.6/45.3   \\
18 & Vote-$k$~\cite{su2022selective} &ICLR 2023 & 61.1/67.2/52.7  & 39.5/41.0/37.9 & 85.5/89.5/78.9 & 57.2/50.0/54.7 & 40.0/42.6/33.2  \\
18 & IDEAL~\cite{zhang2023ideal}  & ICLR 2024  & 65.1/66.4/62.5  & 39.5/46.1/35.9  & 78.0/89.5/59.8 & 57.3/63.7/50.8 & 45.8/50.8/40.6 \\
\rowcolor{gray!50}
18 & Sub-SA &  Ours & \textbf{65.9/68.4/64.1}  & \textbf{40.4/42.6/39.1}  & \textbf{90.6/93.4/88.7}& \textbf{61.5/62.5/59.8}& \textbf{49.7/55.5/42.2}  \\
\hline
\end{tabular}%
}
\caption{Mean/Maximum/Minimum evaluation results of all methods, Comparison on five different datasets for the classification task with an annotation budget of 100 and 18. We apply similar-based prompt retrieval to get the $k$ shot examples for all methods. The average results are obtained from three different runs for each model. Our method shows better performance than Random, Vote-$k$, and IDEAL in almost all cases. The best result in each case is \textbf{bolded}.}
\label{con:t1}
\end{table*}
%%%%%%%%%%%%%%%%%%%%%%%%%%%%%%%%%%%%%%%%%%%%%%%%%%%%%%%%%%%%%%%%%%%%%%%%% 
\section{Experiments}
\vspace{10pt} % 增加 12 磅的垂直间距
\label{Experiments}
In this section, we perform experiments across eight different NLP datasets with diverse categories of tasks (including classification, commonsense reasoning, dialogue, and text generation).
We first introduced the experimental setups and implementation details in Section~\ref{sub: Experimental Setups} and Section~\ref{sub: Implementation Details}. Then, in Section~\ref{sub: Main Results}, we present the proposed method Sub-SA, which can effectively and efficiently ascertain better selective annotation compared with the baselines. Finally, we demonstrate the excellent performance of the proposed method in three different settings (alternative selective annotation
methods, various retrieval approaches, and different models) for comprehensive comparison in Section~\ref{sub: Analysis}.

\begin{table}
\centering
\resizebox{\columnwidth}{!}{%
\begin{tabular}{llcc}
\hline & Dataset & Task & In-Context Learning Models \\
\hline & MRPC~\cite{dolan2004unsupervised}  & Paraphrase Detection &  GPT-J \\
       & MNLI~\cite{williams2017broad}  & Natural Language Inference & GPT-Neo, GPT-J \\
Classification& SST-2~\cite{socher2013recursive} & Sentiment Analysis & GPT-J \\
       & RTE~\cite{bentivogli2009fifth}  & Natural Language Inference & GPT-Neo, GPT-J \\
       & SST-5~\cite{socher2013recursive} & Sentiment Analysis & GPT-J \\
\hline Multiple-Choice & HellaSwag~\cite{zellers2019hellaswag}  & Commonsense Reasoning &  GPT-J \\
\hline Dialogue & MWoZ 2.4~\cite{budzianowski2018multiwoz}  & Dialogue State Tracking &  GPT-3  \\
\hline Generation  & GeoQuery~\cite{zelle1996learning} & Semantic Parsing & GPT-3.5-Turbo \\
\hline
\end{tabular}}
\caption{All of the eight datasets with examples used in our experiments. These datasets cover various domains, including classification, multiple-choice selection, dialogue, and generation. We also present the in-context learning models for dataset evaluation. GPT-J and GPT-3.5-Turbo are used by default. GPT-Neo is explored in the analysis.}
\vspace{10pt} % 增加 12 磅的垂直间距
\label{con:dataset}
\end{table}

\begin{table*}
\centering
\resizebox{0.6\textwidth}{!}{%
\begin{tabular}{cccccccccccccccccccccccc} % 将所有的 'l' 改为 'c' 来实现居中对齐
\hline
\multicolumn{2}{c}{\textbf{Method}} & \multirow{2}{*}{\textbf{Publication}}  & \multicolumn{1}{c}{\textbf{Multi-Choice}} & \multicolumn{1}{c}{\textbf{Dialogue}} & \multicolumn{1}{c}{\textbf{Generation}}  \\
\cmidrule(lr){1-2} \cmidrule(lr){4-6} 
\multirow{2}{*}{\textbf{$|\mathcal{A}|$}} & \multirow{2}{*}{\textbf{Selection}} & &  HellaSwag & MWoZ & GeoQ   \\ 
\cmidrule(lr){4-6} 
& & & \scriptsize{Mean/Max/Min} & \scriptsize{Mean/Max/Min} & \scriptsize{Mean/Max/Min}  \\
\hline
100 & Random  & -  & 66.9/70.3/64.5 & 40.5/44.5/37.1 & 57.9/58.6/57.0 \\
100 & Vote-$k$~\cite{su2022selective}  & ICLR 2023   & 66.9/69.9/65.2  & \textbf{48.6/49.2/47.7} & 59.4/60.9/57.0   \\
100 & IDEAL~\cite{zhang2023ideal}  & ICLR 2024 & 66.8/69.5/64.8 & 43.2/52.3/38.3 & 60.3/60.9/59.0  \\
\rowcolor{gray!50}
100 & Sub-SA & Ours   & \textbf{67.7/69.5/64.8} & 43.6/44.1/43.4 & \textbf{60.4/61.7/59.4} \\
\hline
18 & Random  & -  & 66.0/68.4/64.1 & 36.7/44.1/28.9 & 48.4/50.0/46.9   \\
18 & Vote-$k$~\cite{su2022selective} &ICLR 2023  & 65.9/68.8/64.1 & \textbf{44.5/45.7/43.0} & 50.1/51.2/48.8 \\
18 & IDEAL~\cite{zhang2023ideal}  & ICLR 2024   & 65.4/70.3/62.5 & 36.2/41.8/25.8 & 44.3/46.9/42.6  \\
\rowcolor{gray!50}
18 & Sub-SA &  Ours   &  \textbf{66.0/69.5/63.7} & 40.5/46.1/36.3& \textbf{53.9/55.1/53.1} \\
\hline
\end{tabular}%
}
\caption{Mean/Maximum/Minimum evaluation results of all methods, Comparison on different datasets for the multiple-choice, dialogue, and generation tasks with an annotation budget of 100 and 18. We apply similar-based prompt retrieval to get the $k$ shot examples for all methods. The average results are obtained from three different runs for each model. Our method shows better performance than Random, Vote-$k$, and IDEAL in multiple-choice and generation tasks. The best result in each case is \textbf{bolded}.}
\vspace{10pt}
\label{con:t2-t4}
\end{table*}

\subsection{Experimental Setups}
\label{sub: Experimental Setups}
\noindent\textbf{\noindent\textbf{Experimental Datasets.}}
For extensive evaluations, we select eight diverse datasets from different NLP tasks, including classification, commonsense reasoning, dialogue, and generation. Table~\ref{con:dataset} describes details of the datasets.
Following previous work~\cite{su2022selective,zhang2023ideal}, we split the "train/dev/test" dataset from the Transformer library~\cite{wolf2019huggingface}, and apply test data for evaluation in SST-5, SST-2~\cite{socher2013recursive}, MWoZ~\cite{budzianowski2018multiwoz} datasets. 
For the rest, we use the development dataset to evaluate the performance as former research~\cite{su2022selective,zhang2023ideal}.
We evaluate the method by accuracy for classifications and multiple choices tasks;
for dialogue task MWoz~\cite{budzianowski2018multiwoz}, we apply joint accuracy~\cite{budzianowski2018multiwoz}; And for generation task GeoQuery~\cite{zelle1996learning}, we adopt the test suite accuracy~\cite{zhong2020semantic}.

\vspace{10pt} % 增加 12 磅的垂直间距
% \subsubsection{ Models }
\noindent\textbf{\noindent\textbf{Base Models.}}
In our study, we present the main empirical results by using the GPT-J with 6B parameters~\cite{gpt-j}, except the GeoQuery and MWoZ datasets, which we apply GPT-3.5-Turbo with 175B parameters ~\cite{brown2020language} to measure the performance. We also adopt GPT-Neo with 2.7B parameters ~\cite{black_2021_5551208} to estimate the robustness of proposed methods. Unless otherwise stated, all experiments are conducted with GPT-J (6B) model.

\vspace{10pt} % 增加 12 磅的垂直间距
\noindent\textbf{\noindent\textbf{Baseline Methods.}}
% \subsubsection{Baselines}
We compare the proposed Sub-SA with three baselines: Random selection, Vote-$k$~\cite{su2022selective}, and IDEAL~\cite{zhang2023ideal}. 
A simple baseline method is the random selection, which does not involve any analysis. This approach randomly selects a series of specified examples from the unlabeled data pool. 
Vote-$k$~\cite{su2022selective} first selects small samples to annotate, then prediction the other unlabeled data with these small set examples. Finally, Vote-$k$ selects the high confidence score examples based on the prediction results. Figure~\ref{fig: Methods} (a) shows the framework of Vote-$k$.
IDEAL~\cite{zhang2023ideal} adopts an influence-driven mechanism to select examples from a directed graph that is constructed with the unlabeled data pool. Figure~\ref{fig: Methods} (b) shows the architecture of IDEAL. To ensure fairness, we use the official code to conduct experiments\footnote{\url{https://github.com/skzhang1/IDEAL/}} of  Vote-$k$ and IDEAL and compare with Sub-SA under the same hardware condition.
 
\subsection{Implementation Details}
\label{sub: Implementation Details}
\noindent\textbf{Details of Getting Unlabeled Data.}
As the previous state-of-the-art selective annotation methods~\cite{su2022selective,zhang2023ideal}, we simulate the realistic setting to obtain unlabeled examples. Specifically, we conduct selective annotation from 3K instances that are randomly taken from the original training data for each task. For each experiment, we repeat this subsampling procedure three times, and average the results over the three trials. Table~\ref{con:t1} and~\ref{con:t2-t4} present the mean/maximum/minimum evaluation results from these three trials. We will find that Sub-SA demonstrates empirical success across previous state-of-the-art selective annotation methods.

\vspace{10pt} % 增加 12 磅的垂直间距
\noindent\textbf{Experimental Conditions.}
We apply PyTorch~\cite{paszke2019pytorch} to implement our method and the baselines. For GPT-3.5-Turbo, we perform the experiments by calling the OpenAI API using a MacBook. The GPT-J 6B and GPT-Neo 2.7B models are from the Huggingface transformer library~\cite{wolf2019huggingface}. We experiment with three different random seeds and present the mean, maximum, and minimum outcomes. All our experiments of GPT-J 6B and GPT-Neo 2.7B  are conducted on a single NVIDIA A40 (48GB) GPU.

\subsection{Main Results}
\label{sub: Main Results}
\begin{table*}[htbp]
\centering
\resizebox{0.9\textwidth}{!}{%
\begin{tabular}{ccccccccccccccccccc} % 将所有的 'l' 改为 'c' 来实现居中对齐
\hline
\multicolumn{2}{c}{\textbf{Method}} & \multicolumn{5}{c}{\textbf{Classification}} & \multicolumn{1}{c}{\textbf{Multi-Choice}} & \multicolumn{1}{c}{\textbf{Dialogue}} & \multicolumn{1}{c}{\textbf{Generation}}  \\
\cmidrule(lr){1-2}\cmidrule(lr){3-7} \cmidrule(lr){8-8} \cmidrule(lr){9-9} \cmidrule(lr){10-10} 
$|\mathcal{A}|$ & Selection & MRPC (s)  & MNLI (s)  & SST-2 (s)& RTE (s) & SST-5 (s) & HellaSwag (s) & MWoZ (s) & GeoQ (s)   \\ 
\hline
100 & Vote-$k$~\cite{su2022selective}  & 4698  & 7051  & 4744  & 3948  & 11760  & 9569  & 15981  & 332  \\
100 & IDEAL~\cite{zhang2023ideal} & 27038  & 26655  & 27174  & 20699  & 26579  & 24694  & 12489  & 1066  \\
\rowcolor{gray!50}
100 & Sub-SA (Ours) & \textbf{0.27 } & \textbf{0.27 } & \textbf{0.32 } & \textbf{0.27} & \textbf{0.29 } & \textbf{0.42 } & \textbf{2.4 } & \textbf{0.02 } \\ 
\hline
100 & {\textbf{Speedup$\uparrow$}}& 17000$\times$ &  26000$\times$ & 14000$\times$& 14000$\times$ & $40000\times$&22000$\times$&$6000\times$&16000$\times$\\
\hline
18 & Vote-$k$~\cite{su2022selective}  & 4748  & 7097  & 4750  & 3955  & 11749  & 9453  & 10844  & 319  \\
18 & IDEAL~\cite{zhang2023ideal} & 824  & 802  & 828  & 593  & 830  & 722  & 379  & 55  \\
\rowcolor{gray!50}
18 & Sub-SA (Ours) & \textbf{0.19 } & \textbf{0.22 } & \textbf{0.24 } & \textbf{0.20 } & \textbf{0.18 } & \textbf{0.17 } & \textbf{2.4 } & \textbf{0.01 } \\
\hline
18 &{\textbf{Speedup$\uparrow$}} & 24000$\times$  & 32000$\times$  & 19000$\times$  & 19000$\times$  & 65000$\times$  & 55000$\times$  & 4000$\times$  & 31000$\times$ \\
\hline
\end{tabular}%
}
\caption{Comparison of time consumption during subset selection under the same hardware condition with an annotation budget of 100 and 18. The results are expressed in seconds. As can be seen, our method presents the shortest time on each dataset and achieves millisecond-level selection time for in-context learning. The best result in each case is \textbf{bolded}. }
\vspace{10pt}
\label{con: time consumption}
\end{table*}

\definecolor{color1}{RGB}{214, 39, 40} % Random
\definecolor{color2}{RGB}{31, 119, 180} % Vote-k
\definecolor{color3}{RGB}{44, 160, 44} % IDEAL
\definecolor{color4}{RGB}{255, 165, 0} % Vote
\begin{figure*}[!h]
\centering
\resizebox{\textwidth}{!}{
\begin{minipage}{\columnwidth}
\centering
\begin{tikzpicture}
\begin{axis}[
  ybar,
  % log basis y={10}, % 使用对数刻度
  ymin=0, ymax=10, % 调整对数刻度的范围
  tick label style={font=\scriptsize},
  ylabel={ $\textbf{Log}$ Time (ms)},
  ylabel style={font=\scriptsize}, % 调整 y 轴标签的字体大小
  symbolic x coords={MRPC, MNLI, SST-2, RTE, SST-5, HeSwag, MWoZ, GeoQ},
  xtick=data,
  xticklabels={ MRPC, MNLI, SST-2, RTE, SST-5, HeSwag, MWoZ, GeoQ},
  xticklabel style={rotate=0, anchor=east, font=\scriptsize, yshift=-5pt, xshift=15pt}, % 向下移动 x 轴标签
  xtick style={draw=none},
  width=1\textwidth, % 调整图形宽度以更好地适应
  height=0.5\textwidth, % 调整图形高度以更好地适应
  bar width=6pt,
  ymajorgrids=true,
  enlarge x limits=0.07, % 调整此值来缩小数据之间的距离
  legend style={at={(0.99,0.99)}, 
    anchor=north east, 
    legend columns=-1,
    font=\tiny},
  grid=none,
  xtick style={draw=none},
  ytick style={draw=none},
]
\legend{Vote-$k$,IDEAL,Sub-SA}
\addplot[fill=color2,legend image code/.code={\fill[color2] (0cm,-0.1cm) rectangle (0.3cm,0.1cm);}] 
coordinates {(MRPC,6.672) (MNLI,6.848) (SST-2,6.676) (RTE,6.596) (SST-5,7.070) (HeSwag,6.981) (MWoZ,7.204) (GeoQ,5.521)};
\addplot[fill=color3, legend image code/.code={\fill[color3] (0cm,-0.1cm) rectangle (0.3cm,0.1cm);}] 
coordinates {(MRPC,7.432) (MNLI,7.426) (SST-2,7.434) (RTE,7.316) (SST-5,7.425)  (HeSwag,7.393) (MWoZ,7.097) (GeoQ,6.028)};
\addplot[fill=color1, legend image code/.code={\fill[color1] (0cm,-0.1cm) rectangle (0.3cm,0.1cm);}] 
coordinates {(MRPC,2.431) (MNLI,2.431) (SST-2,2.505) (RTE,2.432) (SST-5,2.462)  (HeSwag,2.623) (MWoZ,3.380) (GeoQ,1.302)};
\end{axis}
\end{tikzpicture} 
\caption*{\scriptsize a. Annotation Budget is 100} 
\end{minipage}%

\begin{minipage}{\columnwidth}
\centering
\begin{tikzpicture}
\begin{axis}[
  ybar,
  % log basis y={10}, % 使用对数刻度
  ymin=0, ymax=10, % 调整对数刻度的范围
  tick label style={font=\scriptsize},
  ylabel={ $\textbf{Log}$ Time (ms)},
  ylabel style={font=\scriptsize}, % 调整 y 轴标签的字体大小
  symbolic x coords={MRPC, MNLI, SST-2, RTE, SST-5, HeSwag, MWoZ, GeoQ},
  xtick=data,
  xticklabels={ MRPC, MNLI, SST-2, RTE, SST-5, HeSwag, MWoZ, GeoQ},
  xticklabel style={rotate=0, anchor=east, font=\scriptsize, yshift=-5pt, xshift=15pt}, % 向下移动 x 轴标签
  xtick style={draw=none},
  width=1\textwidth, % 调整图形宽度以更好地适应
  height=0.5\textwidth, % 调整图形高度以更好地适应
  bar width=6pt,
  ymajorgrids=true,
  enlarge x limits=0.07, % 调整此值来缩小数据之间的距离
  legend style={at={(0.99,0.99)}, 
    anchor=north east, 
    legend columns=-1,
    font=\tiny},
  grid=none,
  xtick style={draw=none},
  ytick style={draw=none},
]
\legend{Vote-$k$,IDEAL,Sub-SA}
\addplot[fill=color2,legend image code/.code={\fill[color2] (0cm,-0.1cm) rectangle (0.3cm,0.1cm);}] 
coordinates {(MRPC,6.676) (MNLI,6.851) (SST-2,6.677) (RTE,6.597) (SST-5,7.07) (HeSwag,6.976) (MWoZ,7.035) (GeoQ,5.504)};
\addplot[fill=color3, legend image code/.code={\fill[color3] (0cm,-0.1cm) rectangle (0.3cm,0.1cm);}] 
coordinates {(MRPC,5.916) (MNLI,5.904) (SST-2,5.918) (RTE,5.773) (SST-5,5.919)  (HeSwag,5.859) (MWoZ,5.579) (GeoQ,4.74)};
\addplot[fill=color1, legend image code/.code={\fill[color1] (0cm,-0.1cm) rectangle (0.3cm,0.1cm);}] 
coordinates {(MRPC,2.279) (MNLI,2.342) (SST-2,2.38) (RTE,2.30) (SST-5,2.255)  (HeSwag,2.230) (MWoZ,3.380) (GeoQ,1)};
\end{axis}
\end{tikzpicture}
\caption*{\scriptsize b. Annotation Budget is 18} 
\end{minipage}}
\vspace{10pt}
\caption{Comparison of Vote-$k$, IDEAL, and SubSA with respect to time consumption for subset selection under the same hardware condition. The y-axis represents the $\textbf{log}$ presentation with a millisecond scale under 100/18 annotation budget. For the  $\textbf{log}$ millisecond level, our SubSA achieves the lowest selection time for In-context learning.}
\label{fig: 10018}
\end{figure*}

\noindent\textbf{Effectiveness.} We present the performance on eight datasets, covering classification, commonsense reasoning, dialogue, and generation tasks, with Random, Vote-$k$, IDEAL, and our method Sub-SA. Note that for the sake of fairness, we select 18 and 100 as the annotation budget respectively.  Following the experimental conditions setting, the result is reported in Table~\ref{con:t1} and Table~\ref{con:t2-t4}.  From the results, we have the following findings: First, our method outperforms than the baselines in the evaluation tasks on 13 out of 16, especially in classification tasks. This improvement is evidence the Sub-SA effectively selects the annotation across most NLP datasets; Second, in-context learning with 18 examples selected by the Sub-SA achieves higher performance than the one with 100 randomly selected constructors on 6 out of 8 tasks. The deterministic selective annotation method Sub-SA shows stability and capability of better in-context learner; Last, as the OpenAI deprecated the Codex-davinci-002 model, we apply the GPT-3.5-Turbo API to evaluate the dialogue and generation tasks. For the MWoZ task, the Vote-$k$ shows better performance than others, these results present the robust ability of the GPT-3.5-Turbo model on prompt selection for dialogue task. 

\vspace{10pt}
\noindent\textbf{Efficiency.} We compare the time consumption of subset selection in our method Sub-SA against Vote-$k$ and IDEAL on all tasks with the same hardware condition. Random selection does not use an optimization analysis, letting this approach randomly select the examples without delay. Therefore, we did not compare Random selection with other methods. As shown in Table~\ref{con: time consumption},
With respect to the LLM's feedback, Vote-$k$ selects the examples based on the diversity confidence score of the prediction for most unlabeled data. This process ensures that different annotation budgets (e.g., 18/100) have the same selection time.
Constructing from the influence-driven mechanism, the selection time of IDEAL increases continuously as the annotation budget increases. Based on the table, the selection time on the minute or even hourly level does not align with real-world use cases. In our work, Sub-SA achieves millisecond-level time consumption during subset selection, which is efficient in realistic scenarios. In Figure~\ref{fig: 10018}, we compare the time cost of subset selection with log presentation under millisecond scale, our SubSA achieves the lowest selection time for In-context learning. Therefore, we recommend that researchers and practitioners use the practical selective annotation method (e.g., Sub-SA) for ICL.

%%%%%%%%%%%%%%%%%%%%%%%%%%%%%%%%%%%%%%%%%%%%%%%%%%%%%%%%%%%%%%%%%%%%%%%%% 
\subsection{Analysis}
\label{sub: Analysis}

\noindent\textbf{Comparisons With Alternative Methods.} For in-depth comparison, We also explore three alternative selective annotation methods from large-scale unlabeled data:
(1) Maximizing facility location (MFL), which aims at optimizing the representativeness of selected subset.
(2) Fast Vote-$k$, which picks $T$ samples with the largest Vote-$k$ scores, and avoids using the pre-trained language model to the computer confidence score for each instance, resulting in a significantly faster process.
(3) Diversity, which focuses on maximizing the diversity of the embeddings for selected examples. We present the results on MRPC, SST-5, and SST-2 tasks, the results shown in Table~\ref{con:result3}. Performance is also averaged over three random trials. As can be seen, Sub-SA consistently outperforms all the other methods, demonstrating its predominance in selective annotation.

\begin{table}
\centering
\resizebox{\columnwidth}{!}{% 调整表格宽度以适应列宽
\begin{tabular}{cccccccc} % 将所有的 'l' 改为 'c' 来实现居中对齐
\hline
Method & MFL& Fast Vote-$k$ & Diversity &Vote-$k$ &IDEAL& Sub-SA\\ 
\hline
MRPC  & 57.2 & 56.9& 63.4 & 61.1  & 65.1 & \textbf{65.9} \\
% MNLI &  40.1 & 57.2 & 37.2 & 38.4  & 39.8 &  42.3\\
SST-5  & 42.4 & 41.1& 45.2 & 40.0 & 45.8 & \textbf{49.7}  \\
% SST-5 & 41.3 & 44.3& 44.5 & 40.0 & 46.6 & \textbf{51.3}  \\
SST-2 & 88.8 &  85.8 & 83.6 & 85.5 & 78.0 & \textbf{90.6}  \\
\hline
\end{tabular}}
\caption{Comparisons of alternative methods that can gain a subset from large-scale unlabeled data. Here the annotation budget is 18. We apply similar-based prompt retrieval for all methods and the results are obtained from three different runs. As can be observed, our method consistently achieves the best performance. The best result in each case is \textbf{bolded}.}
\label{con:result3}
\end{table}

\vspace{10pt} % 增加 12 磅的垂直间距
\noindent\textbf{Evaluation With Different Retrieval Methods.} Up to this point, we have conducted similarity-based prompt retrieval methods. In this part, we use experiments to quantify the effect of the random baseline for prompt retrieval with a 100 annotation budget. 
Table~\ref{con:result4} presents the results. Based on the table, we observe that Vote-$k$, IDEAL, and Sub-SA suffer from inferior performance when the prompt retrieval method shifts from similarity-based to random selection.
Notice that Sub-SA consistently maintains better performance, whether under random selection or similarity-based retrieval methods. Therefore, Sub-SA can 
create a more stable training subset for ICL.

\begin{table}[!h]
\centering
\resizebox{0.8\columnwidth}{!}{
\begin{tabular}{ccccc} 
\hline
 \multicolumn{2}{c}{\textbf{Method}} & \multicolumn{3}{c}{\textbf{Datasets}}   \\
\cmidrule(lr){1-2} \cmidrule(lr){3-5} 
Selection & Retrieval &RTE &MNLI & HellaSwag  \\ 
\hline
Vote-$k$ & Similar & 57.6 & 39.3 &66.9 \\
IDEAL & Similar & 57.3& 39.7 & 66.8  \\
\rowcolor{gray!50}
Sub-SA(Ours) & Similar & \textbf{58.1} & \textbf{43.2} &  \textbf{67.7} \\
\hline
Vote-$k$ &Random & 55.3 & 38.0 &66.1 \\
IDEAL& Random & 55.0 & 37.9 & 65.8 \\
\rowcolor{gray!50}
Sub-SA(Ours)& Random & \textbf{56.9} & \textbf{42.3} & \textbf{66.9}\\
\hline
\end{tabular}}
\caption{Comparison of random and similar prompt retrieval with Vote-$k$, IDEAL, and Sub-SA on RTE, MNLI, and HellaSwag datasets. Here the annotation budget is 100. As can be observed, our method consistently achieves the best performance with different prompt retrieval methods. The best result in each case is \textbf{bolded}.}
\label{con:result4}
\end{table}

% \vspace{10pt} % 增加 12 磅的垂直间距
\noindent\textbf{Measurement with Other Model.} Here we evaluate Sub-SA on GPT-Neo 2.7B model. The evaluation results are presented in Figure~\ref{fig:result5} with three different random trials. Overall, our evaluation reveals that Sub-SA continuously exceeds the baselines.
Notably, we notice that all the baselines achieve better performance in the GPT-J 6B model than the GPT-Neo 2.7B model. Such results indicate that the size and number of parameters of the model are important factors affecting its performance.

% \vspace{40pt}
\definecolor{color1}{RGB}{214, 39, 40} % Random
\definecolor{color2}{RGB}{31, 119, 180} % Vote-k
\definecolor{color3}{RGB}{44, 160, 44} % IDEAL
\definecolor{color4}{RGB}{255, 165, 0} % Vote
\begin{figure} % Use figure* for two-column wide figures
\centering
\resizebox{\columnwidth}{!}{%
\begin{minipage}{.32\textwidth}
\begin{tikzpicture}[scale=0.65]
\begin{axis}[
ybar,
enlarge x limits=0.5,
legend style={at={(0.98,0.98)}, 
    anchor=north east, 
legend columns=-1,
font=\scriptsize},
ylabel={Performance (\%)},
symbolic x coords={GPT-Neo,GPT-J},
xtick=data,
nodes near coords,
nodes near coords align={vertical},
nodes near coords style={font=\scriptsize}, % Set the font size to tiny for the nodes near coords
ymin=0,ymax=55,
bar width=20pt,
xtick style={draw=none},
ytick style={draw=none},
grid=none,
]
\addplot[fill=color4,legend image code/.code={\fill[color4] (0cm,-0.1cm) rectangle (0.3cm,0.1cm);}] coordinates {(GPT-Neo,34.4) (GPT-J,38.3) };
\addplot[fill=color2,legend image code/.code={\fill[color2] (0cm,-0.1cm) rectangle (0.3cm,0.1cm);}] coordinates {(GPT-Neo,35.4) (GPT-J,39.5) };
\addplot[fill=color3,legend image code/.code={\fill[color3] (0cm,-0.1cm) rectangle (0.3cm,0.1cm);}] coordinates {(GPT-Neo,34.5) (GPT-J,39.5) };
\addplot[fill=color1, legend image code/.code={\fill[color1] (0cm,-0.1cm) rectangle (0.3cm,0.1cm);}] coordinates {(GPT-Neo,35.7) (GPT-J,40.4) };
\legend{Random,Vote-$k$,IDEAL,Sub-SA}
\end{axis}
\end{tikzpicture}
\caption*{\scriptsize a. MNLI} 
\end{minipage}%
\hfill
\begin{minipage}{.32\textwidth}
\begin{tikzpicture}[scale=0.65]
\begin{axis}[
ybar,
enlarge x limits=0.5,
legend style={at={(0.98,0.98)}, 
    anchor=north east, 
legend columns=-1,
font=\scriptsize},
ylabel={Performance (\%)},
symbolic x coords={GPT-Neo,GPT-J},
xtick=data,
nodes near coords,
nodes near coords align={vertical},
nodes near coords style={font=\scriptsize}, % Set the font size to tiny for the nodes near coords
ymin=0,ymax=80,
bar width=20pt,
xtick style={draw=none},
ytick style={draw=none},
grid=none,
]
\addplot[fill=color4,legend image code/.code={\fill[color4] (0cm,-0.1cm) rectangle (0.3cm,0.1cm);}] coordinates {(GPT-Neo,54.4) (GPT-J,56.0) };
\addplot[fill=color2,legend image code/.code={\fill[color2] (0cm,-0.1cm) rectangle (0.3cm,0.1cm);}] coordinates {(GPT-Neo,55.6) (GPT-J,57.2) };
\addplot[fill=color3,legend image code/.code={\fill[color3] (0cm,-0.1cm) rectangle (0.3cm,0.1cm);}] coordinates {(GPT-Neo,55.1) (GPT-J,57.3) };
\addplot[fill=color1, legend image code/.code={\fill[color1] (0cm,-0.1cm) rectangle (0.3cm,0.1cm);}] coordinates {(GPT-Neo,58.3) (GPT-J,61.5) };
\legend{Random,Vote-$k$,IDEAL,Sub-SA}
\end{axis}
\end{tikzpicture}
\caption*{\scriptsize b. RTE} 
\end{minipage}
}
\vspace{10pt}
\caption{Comparisons with another model with the 18 annotation budget. As can be seen, our method consistently achieves the best performance compared with baselines in GPT-Neo (2.7B).}
\vspace{15pt}
\label{fig:result5}
\end{figure}
% \vspace{40pt}

%%%%%%%%%%%%%%%%%%%%%%%%%%%%%%%%%%%%%%%%%%%%%%%%%%%%%%%%%%%%%%%%%%%%%%%%% 
\section{Related Work}
\label{Related Work}
\subsection{In-Context Learning}
In-context learning (ICL)~\cite{brown2020language,liu2021makes,min2021metaicl}, which engages with no parameter updates or fine-tuning for downstream tasks, has exhibited competitiveness across multiple Natural Language Understanding (NLU) and Natural Language Generation (NLG) tasks~\cite{devlin2018bert,dong2022survey,liu2023pre}. 
Generally, prompts are crafted as task/test instructions, which are retrieved from a large-scale annotated dataset with paired input-output examples.
Although ICL led a striking property of LLMs in many fields, recent research~\cite{rubin2021learning,lu2021fantastically} indicates the performance is profoundly dependent on the construct prompts. 
Taking this into account, selecting the optimal in-context learning examples has been crucial to improving the capability of ICL.
Previous works can be mainly classified into the following two categories:
Test example-based methods: these methods~\cite{liu2021makes,gao2020making} aim to retrieve analogous examples for every single test case based on similarity. 
Task-based methods: these methods~\cite{li2023finding,diao2023active,wu2022self} focus on obtaining a set of examples that are suitable for all queries on the same tasks. 

In contrast to selected examples from the annotated dataset, 
selection annotation commits to construct prompts from unlabeled large-scale datasets, which is exceptional cost efficiency and more in line with real-world scenarios. 
Prior approaches like Votk-$k$~\cite{su2022selective} and IDEAL~\cite{zhang2023ideal} proposed LLM estimation and influence-driven mechanism to select prompt examples, respectively. 
In this study, we propose Sub-SA (Submodular Selective Annotation) for ICL, 
Experimental results show that Sub-SA presents better performance than existing selective annotation methods across diverse tasks (including classification, commonsense reasoning, dialogue, and text generation) in an end-to-end manner.
Given its millisecond-level processing time for subset selection, Sub-SA proves to be exceptionally practical for ICL scenarios in real settings.

\subsection{Coreset Selection}
With growing volumes of data in artificial intelligence, the challenges of how to organize and analyze large data let us design efficient approaches to distill the dataset.
Coreset selection aims to select such a small subset of the most informative samples from the given large dataset, 
which dramatically reduces the memory and computational cost for the subsequent action~\cite{chen2024less,kothawade2022prism,guo2022deepcore,iyer2021submodular}. 
Various strategies have been explored to achieve this, ranging from 
geometry-based methods to optimization-based methods and submodularity-based methods.
Geometry-based methods assumed data points close to each other in the feature space with similar properties. So removing these redundant data points can certainly boost efficiency.
Related works include Herding~\cite{welling2009herding}, k-Center-Greedy~\cite{sener2017active} and Moderate Coreset~\cite{xia2022moderate}. 
For optimization-based methods, these approaches are modern the coreset selection as a bilevel optimization problem. Existing works including Glister~\cite{killamsetty2021glister} and Retrieve~\cite{killamsetty2021retrieve}. 
Lastly,
submodularity-based methods, which design a function with submodularity to measure the diversity and information of a large dataset. The works included in this method are Log Determinant, Facility Location, and Graph Cut~\cite{iyer2021submodular}. 
In this work, we propose Sub-SA, which is a submodularity-based method, to select a diverse and
representative subset from a large-scale unlabeled data pool that strengthens ICL for LLM to be better learners.

%%%%%%%%%%%%%%%%%%%%%%%%%%%%%%%%%%%%%%%%%%%%%%%%%%%%%%%%%%%%%%%%%%%%%%%%% 
\section{Conclusions}
\label{Conclusions} 
In this paper, we propose \textbf{Sub-SA}, a submodular-based selective annotation method for LLM to be better learners with well-constructed demonstrations. The design of submodular function and \textbf{RPR} in Sub-SA facilitates efficient and effective subset selection from the large unlabeled dataset across diverse tasks (covering classification, commonsense reasoning, dialogue, and text generation), meanwhile exponentially reducing the time consumption (from hour-level to millisecond-level). Theoretically, we demonstrate that our Sub-SA possesses the characteristics of monotonically and submodularity. Empirically, Sub-SA can improve the performance on a series of benchmarks in an end-to-end manner. The effectiveness and efficiency of Sub-SA offer practical implications for in-context learning in real-world contexts, facilitating more effective language-based tasks. We hope our Sub-SA can give more inspiration to the community in ICL.

%%%%%%%%%%%%%%%%%%%%%%%%%%%%%%%%%%%%%%%%%%%%%%%%%%%%%%%%%%%%%%%%%%%%%%%%

%%% Use this environment to include acknowledgements (optional).
%%% This will be omitted in doubleblind mode.

% \begin{ack}
% We would like to thank all the anonymous reviewers for their constructive feedback. 
% \end{ack}

%%%%%%%%%%%%%%%%%%%%%%%%%%%%%%%%%%%%%%%%%%%%%%%%%%%%%%%%%%%%%%%%%%%%%%%%

%%% Use this command to include your bibliography file.

% \bibliography{mybibfile}
\bibliographystyle{IEEEtran}

\end{document}